\documentclass[letterpaper,10pt,conference]{ieeeconf}  

\IEEEoverridecommandlockouts                              
\usepackage[utf8]{inputenc}
\usepackage[T1]{fontenc}
\usepackage{amsmath,amssymb}
\usepackage{graphicx}
\usepackage[utf8]{inputenc}
\graphicspath{{./figs/}}
\usepackage{color}
\usepackage{bm}
\usepackage{float}
\usepackage{url}
\usepackage{cite}
\usepackage{subcaption}
\usepackage[font=footnotesize]{subcaption}
\usepackage[font=footnotesize]{caption}
\usepackage{xspace}
\usepackage{wrapfig}
\usepackage{multirow,multicol}
\usepackage[table]{xcolor}
\usepackage{flushend}

\definecolor{lightgreen}{rgb}{0.8,1,0.8}
\newcommand{\cmark}{\cellcolor{lightgreen}}%

\DeclareMathOperator*{\argmax}{argmax} 
\DeclareMathOperator*{\argmin}{argmin} 

\def\dplan{dGPMP2\xspace}

\title{\LARGE \bf
Differentiable Gaussian Process Motion Planning
}

\author{\textbf{Mohak Bhardwaj\textsuperscript{1}, Byron Boots\textsuperscript{1}, and Mustafa Mukadam\textsuperscript{2}}\\[2mm]
\textsuperscript{1}University of Washington, \textsuperscript{2}Facebook AI Research
\thanks{
This work was done while all authors were affiliated with the Georgia Institute of Technology.
This work was supported in part by ARL DCIST CRA W911NF-17-2-0181. The authors would also like to thank Alexander Lambert for help with code for generating planning environments.
}}

\begin{document}

\maketitle
\thispagestyle{empty}%
\pagestyle{empty}

\begin{abstract}
Modern trajectory optimization based approaches to motion planning are fast, easy to implement, and effective on a wide range of robotics tasks. However, trajectory optimization algorithms have parameters that are typically set in advance (and rarely discussed in detail). Setting these parameters properly can have a significant impact on the practical performance of the algorithm, sometimes making the difference between finding a feasible plan or failing at the task entirely. We propose a method for leveraging past experience to learn how to automatically adapt the parameters of Gaussian Process Motion Planning (GPMP) algorithms.
Specifically, we propose a differentiable extension to the GPMP2 algorithm, so that it can be trained end-to-end from data. We perform several experiments that validate our algorithm and illustrate the benefits of our proposed learning-based approach to motion planning.
\end{abstract}


\section{Introduction}\label{sec:introduction}

Robot motion planning is a challenging problem, as it requires searching for collision-free paths while satisfying robot and task-related constraints for high-dimensional systems with limited on-board computation. Trajectory optimization is a powerful approach to effectively solving the planning problem and state-of-the-art algorithms can find smooth, collision free trajectories in almost real-time for complex systems such as robot manipulators~\cite{ratliff2009chomp,schulman2013finding,Mukadam-IJRR-18}. Although these approaches are easy to implement and generally applicable to a wide range of tasks, they have certain parameters which can strongly affect their performance in practice. This leads to two major problems:
(i) there is no formal way of setting parameters for a given task and thus requires manual tuning that can be time-consuming and arduous; and
(ii) the planner needs to be re-tuned if the distribution of obstacles in the environment changes significantly, making it brittle in practice, i.e. a planner that works well on one type of environment might completely fail on another.
The above issues need to be addressed in order to create flexible robotic systems that can work seamlessly across a variety of environments and tasks. In order to do so, we ask the following question: can we leverage past experience to learn parameters of the planner in a way that directly improves its expected performance over the distribution of problems it encounters?

\begin{figure}
    \centering
    \includegraphics[trim={0 0 0 0},clip,width=\linewidth]{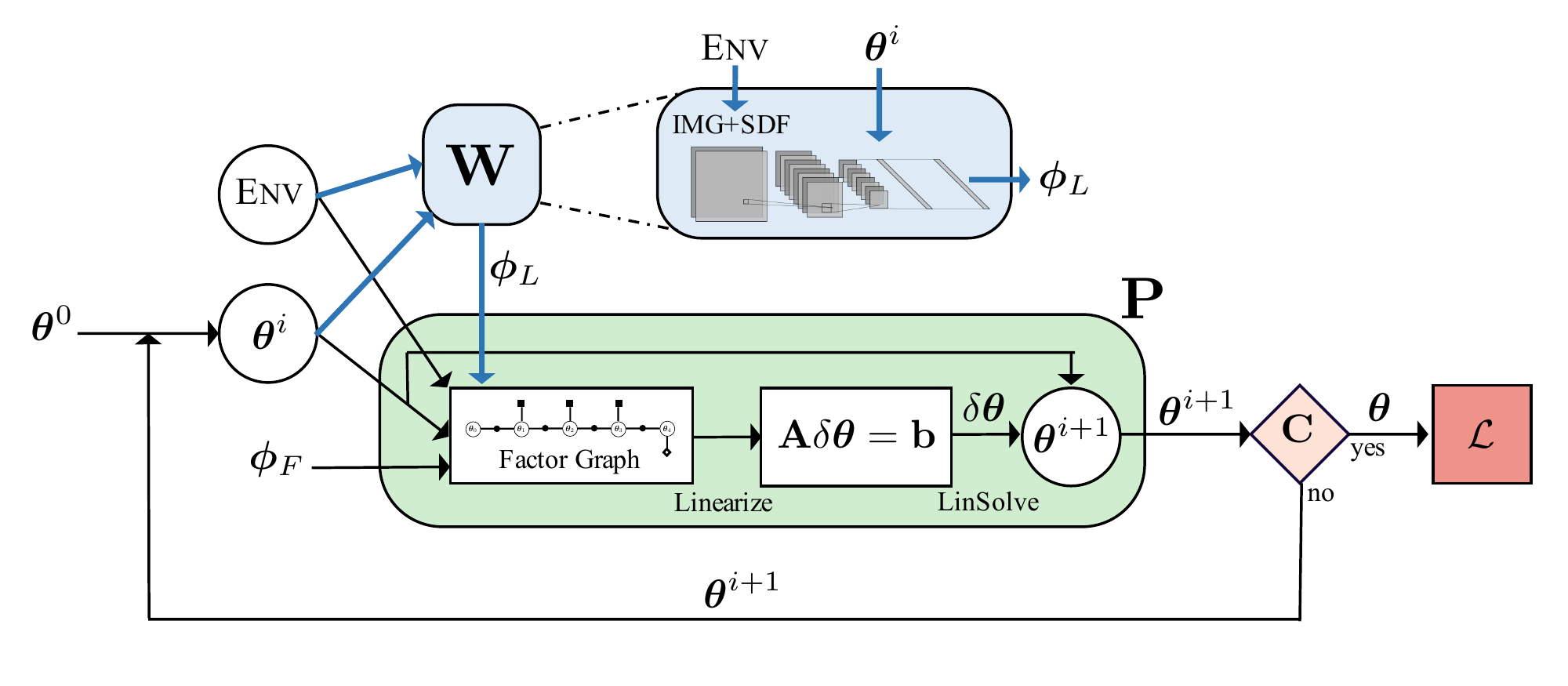}
	\vspace{-6mm}
    \caption{\small
	The computational graph of \dplan where $\bm{\phi}_{F}$ are user defined planning parameters that are fixed and $\bm{\phi}_{L}$ are learned planning parameters. See Section~\ref{sec:approach} for details.}    
	\label{fig:cg}
	\vspace{-6mm}
\end{figure}

In this work, we focus on GPMP2~\cite{Dong-RSS-16}, a state-of-the-art motion planning algorithm that formulates trajectory optimization as inference on a factor graph and finds solutions by solving a nonlinear least squares optimization problem, where the inverse covariances of the factors manifest as weights in the objective function. While GPMP2 has been shown to be a leading optimization-based approach to motion planning~\cite{Mukadam-IJRR-18}, in Section \ref{subsec:sensitivity} we illustrate its sensitivity to its objective function parameters (specifically factor covariances). To contend with this problem, we leverage the key insight that GPMP2 can be rebuilt as a \emph{fully differentiable computational graph} and learn the parameters for its objective function from data in an end-to-end fashion. This allows us to develop a learning strategy that can improve GPMP2's performance on a given distribution of problems. Our differentiable version can be trained in a self-supervised manner or from expert demonstrations to predict covariances that are time and space varying, in contrast to fixed, hand-tuned covariances, as used in the vanilla approach. Building on top of a structured planner offers interpretability and allows us to explicitly incorporate planning constraints such as collision avoidance and velocity limits. 
This work is intended as a preliminary investigation into learning structured planners from high-dimensional inputs. We perform several experiments in simulated 2D environments to demonstrate the benefits of our approach which we call Differentiable GPMP2 (\dplan), illustrated in Fig.~\ref{fig:cg}.

\section{Related Work}\label{sec:related_work}

Machine learning has been used to accelerate motion planning by combining reinforcement learning (RL) with sampling based planning~\cite{faust2018prm}, learning cost functions from demonstration~\cite{ratliff2009learning}, learning efficient heuristics~\cite{bhardwaj2017heuristic}, and learning collision checking policies~\cite{Bhardwaj19a}. As machine learning becomes more accessible, there has been a growing interest in using deep learning for planning such as end-to-end networks to perform value iteration~\cite{tamar2016value} or learning a latent space embedding and a dynamics model that is suitable for planning by gradient descent within a goal directed policy~\cite{srinivas2018universal}. Such approaches have demonstrated that \emph{learning to plan} is a promising research direction as it allows the agent to explicitly reason about future actions. However, learning-based approaches still fall short on several fronts.  Combining learning and planning in a way where domain knowledge, constraints, and uncertainty are properly handled is challenging, and learned representations are often difficult to interpret.

Recent work in structured learning techniques offer avenues towards contending with these challenges. Several methods have focused on incorporating optimization within neural network architectures. For example, \cite{clark2018learning} implicitly learns to perform nonlinear least squares optimization by learning an RNN that predicts its update steps, \cite{NIPS2016_6461} learns to perform gradient descent, and~\cite{chen2018neural} utilizes a ODE solver within its network. Other methods like~\cite{amos2017optnet} learn a sequential quadratic program as a layer in its network, which was later extended to solve model predictive control~\cite{amos2018differentiable}. \cite{byravan2018se3} learns structured dynamics models for reactive visuomotor control. Taking inspiration from this body of work, in this paper we present a differentiable inference-based motion planning technique that through its structure allows us to combine the strengths of both traditional model-based methods and modern learning methods, while mitigating their respective weaknesses.

Another related field of work is on automatic parameter tuning of motion planning algorithms. Approaches such as ~\cite{cano2018automatic, burger2017automated} treat the planner as a black-box and use machine learning tools such as Bayesian optimization, bandits, and random forests to optimize a single configuration of parameters that improves planner performance. However, such a single configuration does not adapt to changes in the environment distribution which hinders generalization. Additionally, the number of parameters optimized in these approaches is far fewer than ours and they do not incorporate high dimensional inputs such as images.

\section{Background}\label{sec:background}

We begin by reviewing the GPMP2~\cite{Mukadam-IJRR-18} planner that we will later reconstruct as a differentiable computational graph. Then, we discuss limitations of GPMP2 with respect to its sensitivity to objective function parameters, thus motivating our learning algorithm.

\subsection{Planning as inference on factor graphs}

We take a probabilistic inference perspective on motion planning as described in the GPMP2 framework~\cite{Dong-RSS-16}. The planning problem is posed as computing the maximum a posteriori (MAP) trajectory given a prior over trajectories and a likelihood of events of interest that the trajectory must satisfy. By selecting appropriate distributions, sparsity can be induced in the MAP problem, which allows for efficient inference. Following~\cite{Dong-RSS-16}, we describe the essential components of GPMP2 here.

\textbf{The Prior:}\quad
In GPMP2, a continuous-time Gaussian process (GP) is used to define a prior distribution over trajectories, $\bm{\theta}(t) \sim \mathcal{GP}(\bm{\mu}(t), \bm{\mathcal{K}}(t, t'))$, where $\bm{\mu}(t)$ is the mean function and $\bm{\mathcal{K}}(t,t')$ is the kernel. For the purposes of our approach, we represent the trajectory using N support states, $\bm{\theta} = [\bm{\theta}_{1}, \ldots, \bm{\theta}_{N}]^T$ at different points in time and define the mean vector and covariance matrix as 
\begin{equation} 
	\bm{\mu} = [\bm{\mu}_1, \ldots, \bm{\mu}_N ]^T \hspace{-2mm}, \quad
	\bm{\mathcal{K}} = [\bm{\mathcal{K}}(t_i, t_j)]\Bigr|_{ij, 1 \leq i,j \leq N}
\end{equation}
and this GP defines a prior on the space of trajectories
\begin{equation}
\label{eq:gauss_prior}
P(\bm{\theta}) \propto \exp \Big\{ -\frac{1}{2} \| \bm{\theta} - \bm{\mu} \|_{\bm{\mathcal{K}}}^{2} \Big\},
\end{equation}
where $\| \bm{\theta} - \bm{\mu} \|_{\bm{\mathcal{K}}}^{2} \doteq (\bm{\theta} - \bm{\mu})^T \bm{\mathcal{K}}^{-1} (\bm{\theta} - \bm{\mu})$ is the Mahalanobis distance.
The GP prior distribution is generated by an LTV-SDE~\cite{barfoot2014batch}
\begin{equation}
\label{eq:ltv_sde}
\dot{\bm\theta} = \bm{A}(t)\bm{\theta}(t) + \bm{B}(t)\bm{u}(t) + \bm{w}(t),
\end{equation}
where $\bm{A}(t)$ and $\bm{B}(t)$ are system matrices, $\bm{u}(t)$ is a bias term, and
$\bm w(t) \propto \mathcal{GP}(\mathbf{0}, \mathbf{Q}_{c}\delta(t - t'))$ is a white noise process with $\mathbf{Q}_c$ being the power spectral density matrix of the system and $\delta$ being the Dirac delta function. The first and second order moments of the solution to Eq.~\eqref{eq:ltv_sde} gives us the mean and covariance of the desired GP prior. The resulting inverse kernel matrix of the GP has an exactly sparse block-tridiagonal structure making it ideal for fast inference. Here, we use the constant velocity prior model, where the covariance for a single time step is specified by
\begin{equation}
\label{eq:const_vel_cov}
\mathbf{Q}_{t_{i}, t_{i+1}} = \begin{bmatrix}
\frac{1}{3} \Delta t_{i}^{3}\mathbf{Q}_c & \frac{1}{2} \Delta t_{i}^{2}\mathbf{Q}_c \\
\frac{1}{2} \Delta t_{i}^{2}\mathbf{Q}_c & \Delta t_{i}\mathbf{Q}_c
\end{bmatrix},
\end{equation}
where $\Delta t_{i} = t_{i+1} - t_{i}$. The full GP covariance is obtained by composing $\mathbf{Q}_{t_{i}, t_{i+1}}$ at every time step along with the start and goal covariances, $\bm{K}_s$ and $\bm{K}_v$. Please refer to~\cite{Dong-RSS-16} and~\cite{barfoot2014batch} for details. One important thing to note here is that the GP prior covariance is completely parameterized by the power spectral density matrix $\mathbf{Q}_{c}$.

\textbf{The Likelihood function:}\quad
The likelihood function is used to capture planning requirements in the form of events $\mathbf{e}$ that the trajectory must satisfy. These include constraints such as collision avoidance, joint or velocity limits, or other task relevant objectives. We define the likelihood function as a distribution in the exponential family given by
\begin{equation}
L(\bm{\theta};\mathbf{e}) \propto \exp \Big\{ -\frac{1}{2} \|\bm{h}(\bm{\theta}) \|_{\bm{\Sigma}}^{2} \Big\},
\end{equation} 
where $\bm{h}(\bm{\theta})$ is a vector-valued cost function and $\mathbf{e}$ are the events of interest. 

\textbf{Inference:}\quad
Given the prior and likelihood, the MAP problem can be solved as
\begin{gather}	
\bm\theta^{*}
= \argmax_{\bm{\theta}} \{ P(\bm{\theta} |\mathbf{e}) \}
= \argmin_{\bm{\theta}} \big\{-\log\big(P(\bm{\theta}) L(\bm{\theta}{;}\mathbf{e}) \big) \big\} \nonumber \\
\bm\theta^{*}
= \argmin_{\bm{\theta}} \Big\{ \frac{1}{2} \| \bm{\theta} -\bm{\mu} \|_{\bm{\mathcal{K}}}^{2} + \frac{1}{2} \|\bm{h}(\bm{\theta}) \|_{\bm{\Sigma}}^{2} \Big\}.  \label{eq:map_prob}
\end{gather}    	     
In general, $\bm{h}(\bm{\theta})$ can be non-linear and thus the above equation is a Nonlinear Least Squares (NLLS) problem which can be solved using iterative approaches like Gauss-Newton or Levenberg-Marquardt (LM) algorithms. At any iteration $i$, these algorithms proceed by first linearizing the cost function around the current estimate of the trajectory, $\bm{\theta}^{i}$, using a Taylor expansion
$\bm{h}(\bm{\theta}) = \bm{h}(\bm{\theta}^{i}) + \mathbf{H} \delta\bm{\theta}$,
where $\mathbf{H} = \frac{\partial \bm{h}}{\partial \bm{\theta}}\Bigr|_{\substack{\bm{\theta}=\bm{\theta}^{i}}}$ and then solving the following linear system to find the update, $\delta \bm{\theta}$:
\begin{equation}
\label{eq:linear_system}
\left( \bm{\mathcal{K}}^{-1} + \mathbf{H}^{T}\bm{\Sigma}^{-1}\mathbf{H} \right)\delta \bm{\theta} = - \bm{\mathcal{K}}^{-1}(\bm{\theta}^{i} - \bm{\mu}) -  \mathbf{H}^{T}\bm{\Sigma}^{-1}\bm{h}(\bm{\theta}^{i}) .
\end{equation}
Gauss-Newton optimization in particular updates the current estimate with the following rule
\begin{equation}
\bm{\theta}^{i+1} = \bm{\theta}^{i} + \delta \bm{\theta}.
\end{equation}
GPMP2 exploits the sparsity of the linear system in Eq.~\eqref{eq:linear_system} to formulate MAP inference on a factor graph and solve it efficiently. While GPMP2 is a state-of-the-art method that outperforms several leading sampling and optimization based approaches to motion planning~\cite{Mukadam-IJRR-18}, it still has some practical limitations with respect to setting the parameters in its objective in Eq.~\eqref{eq:map_prob}. Next, we will discuss these limitations in-depth with a few examples.

\subsection{Sensitivity to objective function parameters} \label{subsec:sensitivity}

The performance of GPMP2 is dependent on the values of $\mathbf{Q}_C$ (the parameter that governs the covariance of the GP prior) and $\mathbf{\Sigma}$ (the covariance of the likelihood) as per its objective function from Eq.~\eqref{eq:map_prob}. For example, for collision avoidance, the distribution of obstacles in the environment affects what relative settings of $\mathbf{Q}_C$ and obstacle covariance $\sigma_{obs}$ (such that $\mathbf{\Sigma} = \sigma_{obs}^2 \times \mathbf{I}$) will be effective in solving the planning problem.
 
\begin{figure}[!t]
\centering
    \begin{subfigure}[t]{0.29\linewidth}
    \centering
    \includegraphics[trim={120 45 120 45},clip,width=\linewidth]{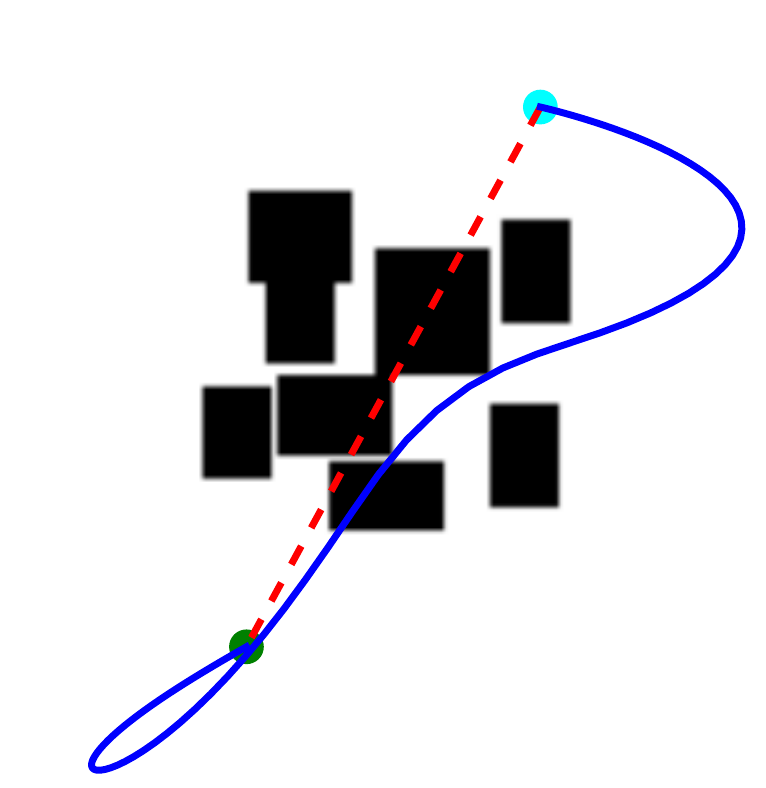}
    \caption{$\sigma_{obs} = 0.1,\\ \bm{Q}_C = 0.5 \times I$}
	\label{fig:tarpit_example1}
    \includegraphics[trim={120 45 120 45},clip,width=\linewidth]{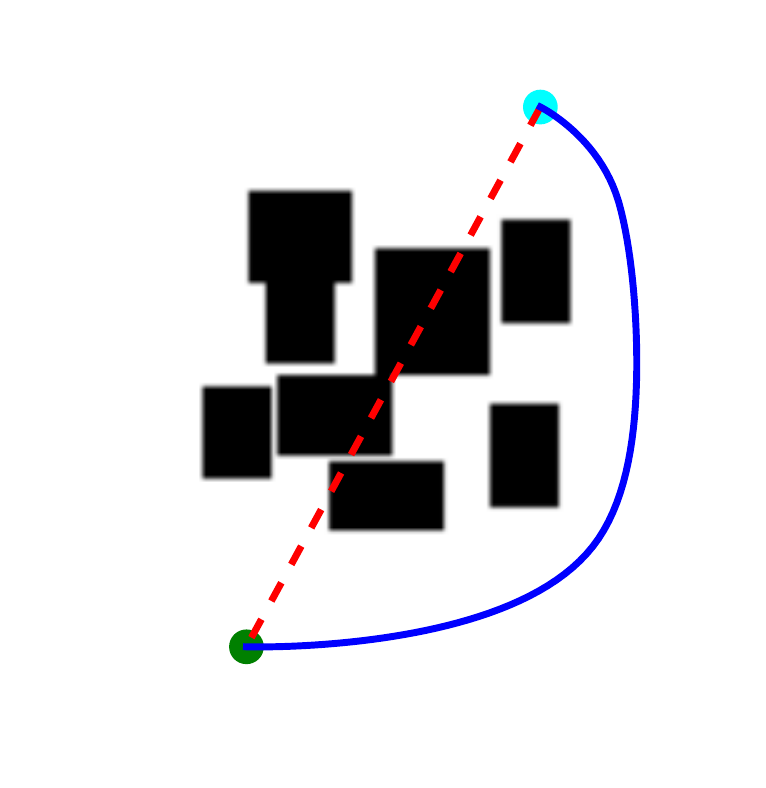}
    \caption{$\sigma_{obs} = 0.01,\\ \bm{Q}_C = 0.5 \times I$}
	\label{fig:tarpit_example2}
    \end{subfigure}
	\hspace{3mm}
    \begin{subfigure}[t]{0.29\linewidth}
    \centering
    \includegraphics[trim={120 45 120 45},clip,width=\linewidth]{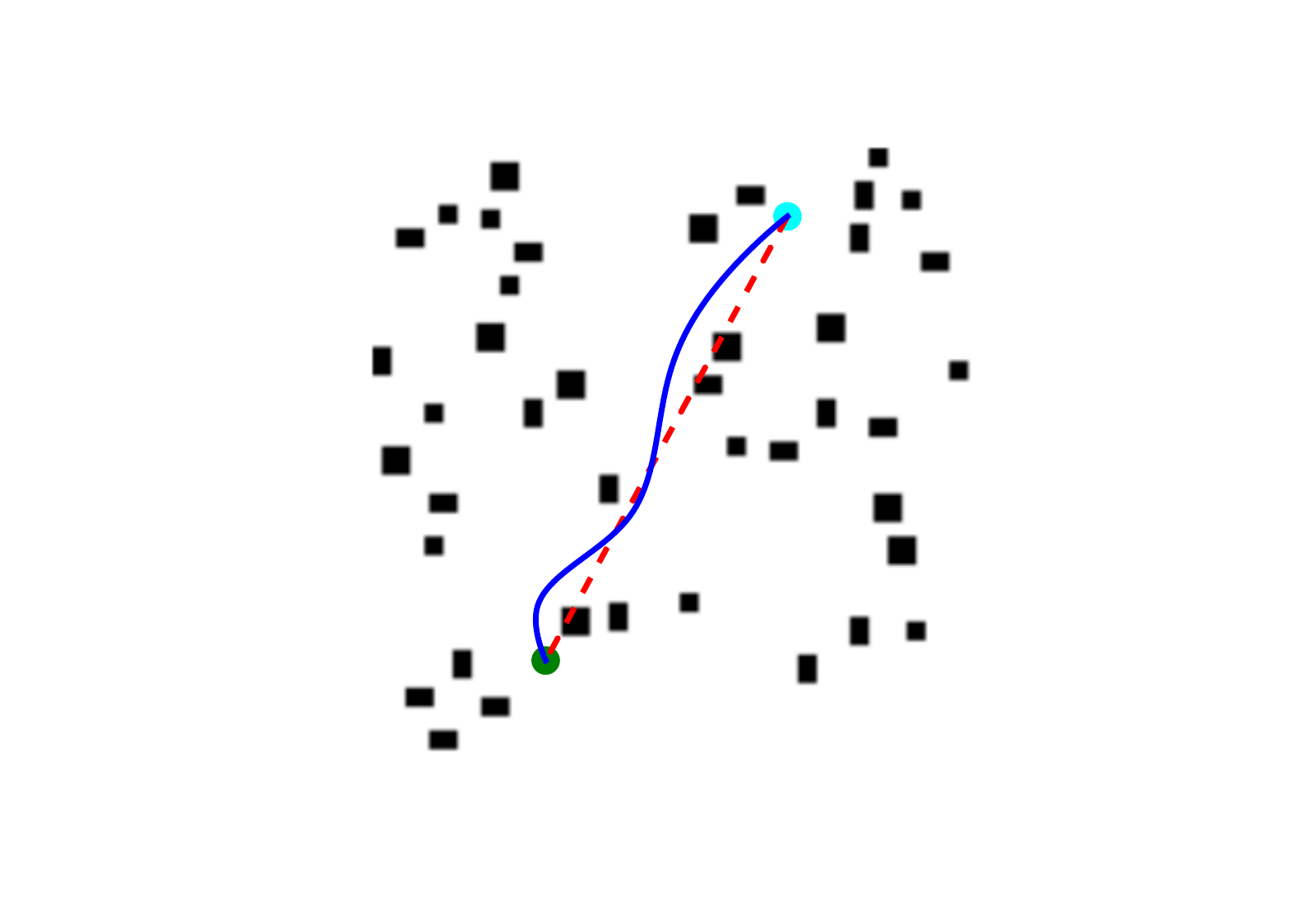}
    \caption{$\sigma_{obs} = 0.1,\\ \bm{Q}_C = 0.5 \times I$}
	\label{fig:forest_example1}
    \includegraphics[trim={120 45 120 45},clip,width=\linewidth]{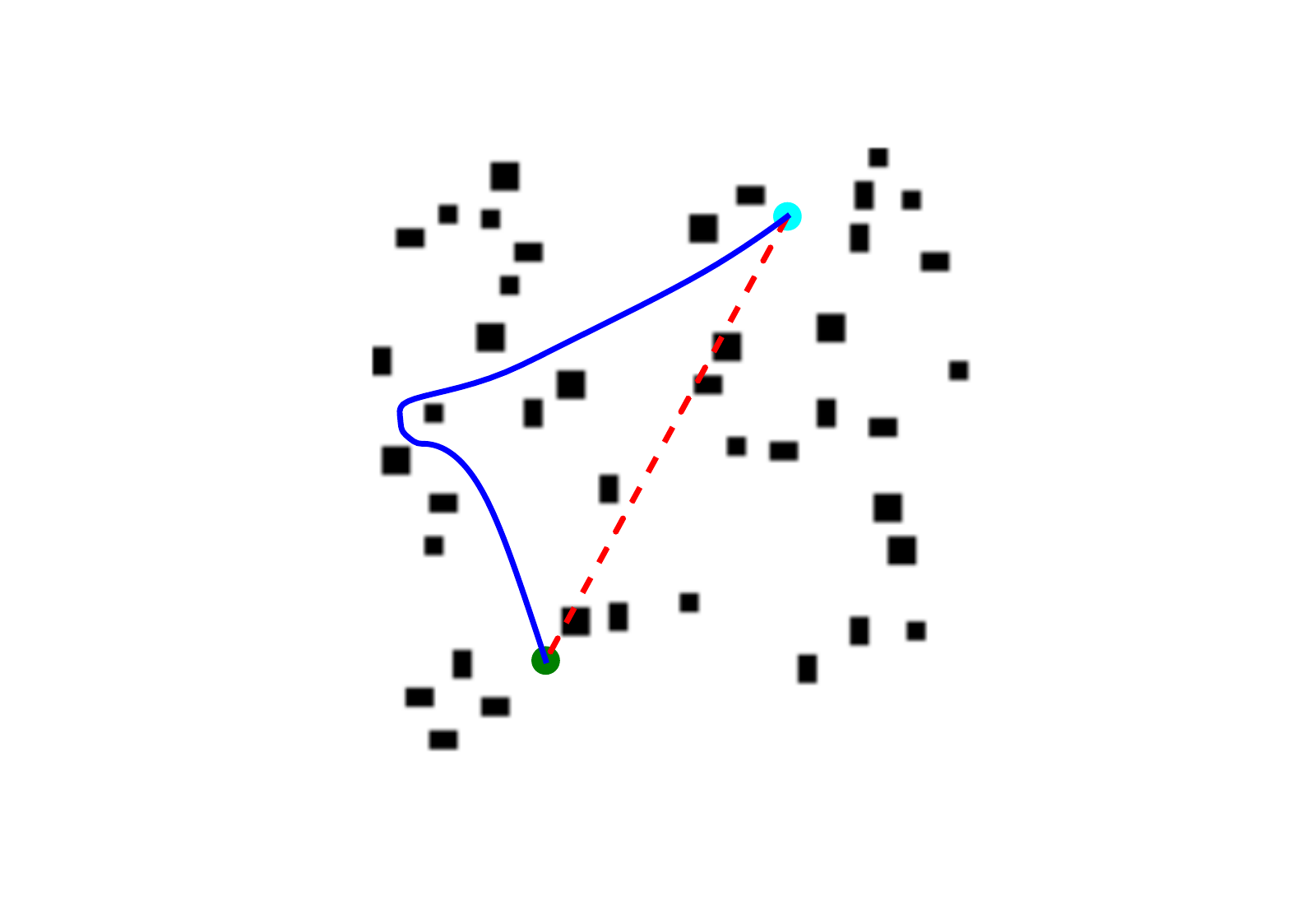}
    \caption{$\sigma_{obs} = 0.01,\\ \bm{Q}_C = 0.5 \times I$}
	\label{fig:forest_example2}
    \end{subfigure}
	\hspace{3mm}
    \begin{subfigure}[t]{0.29\linewidth}
    \centering
    \includegraphics[trim={120 45 120 45},clip,width=\linewidth]{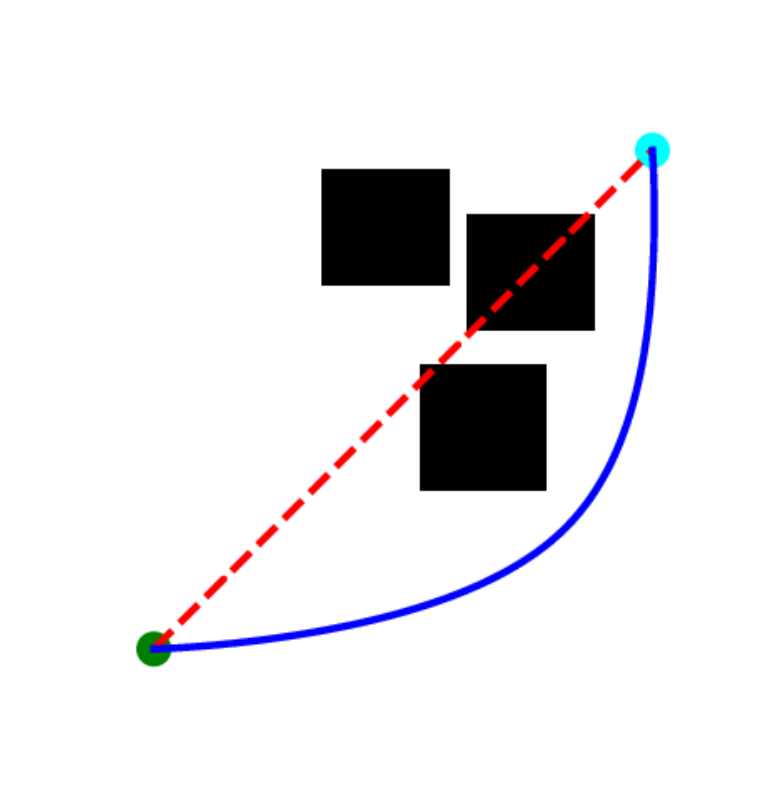}
    \caption{$\sigma_{obs} = 0.03,\\ \bm{Q}_C = I$}
	\label{fig:gap_example1}
    \includegraphics[trim={120 45 120 45},clip,width=\linewidth]{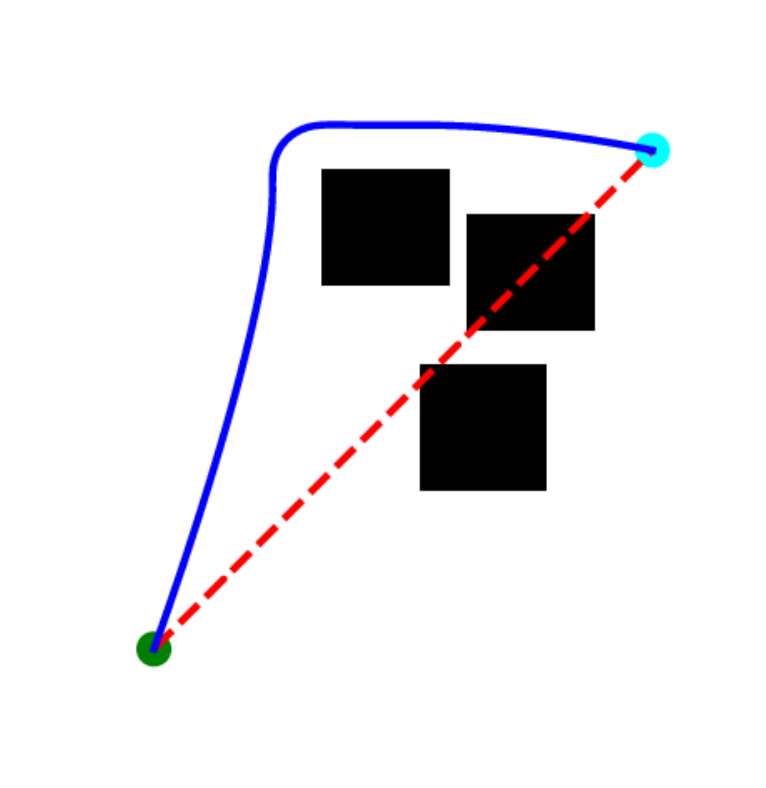}
    \caption{$\sigma_{obs} = 0.01,\\ \bm{Q}_C = I$}%
	\label{fig:gap_example2}
    \end{subfigure}
	\caption{\small
	(a)-(b) $\mathtt{tarpit}$ dataset (robot radius = 0.4m, safety distance = 0.4m). For the same $\bm{Q}_C$, a smaller $\sigma_{obs}$ is required to encourage the planner to navigate around obstacles.
	(c)-(d) $\mathtt{forest}$ dataset (robot radius = 0.2m, safety distance = 0.2m). For the same $\bm{Q}_C$, a larger $\sigma_{obs}$ is required to focus on finding solutions near the straight line trajectory.
	(e)-(f) $\mathtt{multi\_obs}$ dataset (robot radius = 0.4m, safety distance = 0.4m) A small change in obstacle covariance can lead to significant changes in the trajectory.
	In all figures, the red dashed  trajectories are the initializations and the blue trajectories are the optimized solutions.}
    \label{fig:sense}
	\vspace{-5mm}
\end{figure}

Different datasets require different relative settings of parameters.
Due to the nonlinear interactions between these parameters it might not be possible to find a fixed setting that will always work, and in practice it 
can be a tedious task to find a setting that works for many different environments.
For example, in environments like the one in Fig.~\ref{fig:tarpit_example1}-\ref{fig:tarpit_example2}, where the planner needs to find a trajectory that goes around the cluster of obstacles, a small obstacle covariance is required to make the planner navigate around the ``tarpit.'' But, at the same time, if a large dynamics covariance is used, it might try to squeeze in between obstacles where the cost can have a local minima. So a smaller dynamics covariance is needed as well. Another example is shown in Fig.~\ref{fig:forest_example1}-\ref{fig:forest_example2} with dispersed obstacles near the start and goal. Here an entirely different setting of covariances is effective. Since obstacles are small and diffused, solutions can generally be found close to the straight line initialization. A smaller dynamics covariance helps with that. Also, the start and goal can be very near obstacles which means that a small obstacle covariance might lead to solutions that violate the start and goal constraints. Having a smaller obstacle covariance can also lead to trajectories that are very long and convoluted as they try to stay far away from obstacles.

Small changes in parameters can lead to trajectories lying in different homotopy classes. For example, Fig.~\ref{fig:gap_example1}-\ref{fig:gap_example2} illustrates how even minor changes in the obstacle covariance can lead to significant changes in the resulting trajectories. This makes tuning covariances harder, as the effects are further aggravated over large datasets with diverse environments leading to inconsistent results. 

With sufficient domain expertise, the parameters can be hand-tuned. However, this process can be very inefficient and becomes increasingly hard for problems in higher dimensions or when complex constraints are involved. An ideal setup would be to have an algorithm that can predict appropriate parameters automatically for each problem. Therefore, in this work, we rebuild the GPMP2 algorithm as a fully differentiable computational graph, such that these parameters can be specified by deep neural networks which can be trained end-to-end from data. When deployed, our differentiable GPMP2 approach (\dplan) can then automatically select its own parameters given a particular motion planning problem.

\section{A Structured Computational Graph for Motion Planning}\label{sec:approach}

In this section, we first explain how GPMP2 can be interpreted as a differentiable computation graph. Then, we explain how learning can be incorporated in the framework and finally, we show how the entire system can be trained end-to-end from data.

\subsection{Differentiable GPMP2} \label{subsec:dgpmp2}

Our architecture consists of two main components: a planning module $\mathbf{P}$ that is differentiable but has no learnable parameters and a trainable module $\mathbf{W}$ that can be implemented using a differentiable function approximator such as a neural network as shown in Fig.~\ref{fig:cg}. As discussed in Section~\ref{sec:background}, GPMP2 performs trajectory optimization via MAP inference on a factor graph by solving an iterative nonlinear optimization, where at any iteration the factor graph is linearized at the current estimate of the trajectory to produce the linear system in Eq.~\eqref{eq:linear_system} and an update step is computed by solving that linear system. At a high level, our planning module $\mathbf{P}$ implements this update step as a computational graph. The trainable module $\mathbf{W}$ is then set up to parameterize some desired planning parameters and outputs these as $\bm{\phi}_{L}$ at every iteration. These parameters correspond to factor covariances used by $\mathbf{P}$ to construct the linearized factor graph. Additionally, $\mathbf{P}$ takes as input a set of fixed planning parameters $\bm{\phi}_{F}$ to allow parameters that can be user-specified and are not being learned, for example, obstacle safety distance and covariances of constraint factors like start, goal, and velocity. The key insight is that since all operations are differentiable for solving Eq.~\ref{eq:linear_system}, we can easily differentiate through it using standard autograd tools~\cite{paszke2017automatic} and thus train $\mathbf{W}$ in an end-to-end fashion from data.

Similar to GPMP2, during the forward pass, \dplan iteratively optimizes the trajectory where at the $i^{th}$ iteration, the planning module $\mathbf{P}$ takes the current estimate of the trajectory $\bm{\theta}^{i}$ and planning parameters $\bm{\phi}_L$ and $\bm{\phi}_F$ as inputs (where $\bm{\phi}_{L}$ is the output of the trainable module $\mathbf{W}$ and $\phi_{F}$ are user-defined and fixed) and produces the next estimate $\bm{\theta}^{i+1}$ as shown in Fig.~\ref{fig:cg}. The new estimate then becomes the input for the next iteration. This process continues until $\bm{\theta}^{i+1}$ passes a specified convergence check or a maximum of $T$ iterations and the optimization terminates ($\mathbf{C}$). At the end of the optimization, we roll out a complete differentiable computation graph for the motion planner. 

\textbf{Notation:}\quad $\bm{\theta}^{i}$ refers to the trajectory estimate at the $i^{th}$ iteration of the optimization that goes from $1, \ldots, T$ and $\bm{\theta}_{i}$ is the $i^{th}$ state along the trajectory that goes from $1, \ldots, N$.

\textbf{The planning module:}\quad$\bm{\theta}^{i}$ is fed into the planning module along with a signed distance field of the environment and additional planning parameters ($\bm{\phi}_{F}$ and $\bm{\phi}_{L}$) such as factor covariances, safety distance, robot kinematics, start-goal constraints, and other task related constraints. These inputs are used to construct the linear system in Eq.~\eqref{eq:linear_system} corresponding to the linearized factor graph of the planning problem. Similar to standard GPMP2, constraints are implemented as factors with fixed small covariances and the likelihood function for obstacle avoidance is the hinge loss function (see Section~\ref{sec:experiments}) with covariance $\mathbf{\Sigma}$. The trajectory update $\delta\bm{\theta}^{i}$ is then computed by solving this linear system, using Cholesky decomposition of the normal equations~\cite{Mukadam-IJRR-18,dellaert2006square}, and the new trajectory $\bm{\theta}^{i+1}$ is computed using a Gauss-Newton step. The above procedure is fully differentiable and allows computing gradients in the backwards pass with respect to $\bm{\theta}^{i}$, GP covariance $\bm{\mathcal{K}}$, and likelihood function covariance $\bm{\Sigma}$.

\textbf{The trainable module:}\quad The trainable module $\mathbf{W}$ outputs planning parameters $\bm\phi_{L}$. These correspond to covariances of factors in Eq.~\ref{eq:linear_system} that we wish to learn from data. In practice, we can choose to learn the GP covariance $\bm{\mathcal{K}}$, the likelihood covariance $\bm{\Sigma}$, or both. Additionally, this approach allows us to learn individual covariances for different states along the trajectory $[\bm{\theta}_{1}, \ldots, \bm{\theta}_N]$ and different iterations of the optimization thus offering much more expressiveness than a single hand-tuned covariance. We implement $\mathbf{W}$ as a feed-forward convolutional neural network that takes as input the bitmap image of the environment, the signed distance field and the current trajectory $\bm{\theta}^{i}$, and outputs a parameter vector $\bm{\phi}_{L}^{i}$ at every iteration $i$. Note that, given our architecture, $\mathbf{W}$ can be customized as per individual needs based on problem requirements or parameters chosen to be learned.

After a forward pass, we roll out a fully differentiable computation graph that outputs a sequence of trajectories $\{\bm{\theta}^{1}, \ldots, \bm{\theta}^T\}$. Then we evaluate a loss function on this sequence and backpropagate that loss to update the parameters of $\mathbf{W}$ such that it produces parameters $\bm\phi_{L}$ that allow us to optimize for better quality trajectories on the dataset as measured by the loss. We explain our loss function and the training procedure in detail below. 

\subsection{Learning factor graph covariances} \label{sec:learn_cov}

\textbf{Imitation loss:}\quad
Consider the availability of expert demonstrations for a planning problem. These may be provided by an asymptotically optimal (but slow) motion planner~\cite{karaman2011sampling} or by human demonstration~\cite{Rana-CoRL-17}. \dplan can be trained to produce similar trajectories by minimizing an error metric between the demonstrations and learner's output with     
\begin{equation}
\label{eq:loss_imitate}
\mathcal{L}_{imitation} = ||\bm{\theta}^{e} - \bm{\theta}||_{2}^{2} 
\end{equation}
where $\bm{\theta}^{e}$ is the expert's demonstrated trajectory and the metric is the L2 norm. 

\textbf{Task loss:}\quad
Naively trying to match the expert can be problematic for a motion planner.
For example, when equally good paths lie in different homotopy classes, the learner may land in a different one than the expert. In this case, penalizing for not matching the expert may be excessively conservative. If using human demonstrations as an expert, a realizability gap can arise when the planner has different constraints as compared with the human. Thus, we use an external task loss as a regularizer that encourages smoothness and obstacle avoidance, while respecting start and goal constraints, as is often used in motion planning~\cite{zucker2013chomp}:
\begin{equation}
\label{eq:loss_chomp}
\mathcal{L}_{plan} = \mathcal{F}_{smooth} + \lambda \times \mathcal{F}_{obs},
\end{equation}
where $\mathcal{F}_{smooth}$ corresponds to the GP prior error and $\mathcal{F}_{obs}$ is the obstacle cost that are described in Eq.~\eqref{eq:map_prob} and $\lambda$ is a user specified parameter. In practice, the performance is not sensitive to the setting of $\lambda$. Then, the overall loss for a single trajectory is, $\mathcal{L} =  \mathcal{L}_{imitation} + \mathcal{L}_{plan}$.  Note that our framework allows for any choice of loss function depending on the application.

\textbf{Training:}\quad
During training we roll out our learner for a fixed number of iterations $T$ and use Backpropagation Through Time (BPTT)~\cite{rumelhart1988learning} on the sum of losses of the intermediate trajectories in order to update the parameters of the trainable module $\mathbf{W}$. Then, the total loss minimized for our learner over a batch of size $K$ is
\begin{equation}
\mathcal{L}_{total} = \frac{1}{K} \frac{1}{T} \sum_{k=1}^{K}   \sum_{i=1}^{T} \mathcal{L}^{k,i}.
\end{equation}

\section{Experimental Evaluation}\label{sec:experiments}

We test our approach on 2D navigation problems with complex environment distributions and problems with user-specified velocity constraints. Many real world motion planning problems such as warehouse automation (KIVA systems), extra-terrestrial rovers, in-home robots (Roomba), navigation from satellite data, and last mile delivery, among others, are inherently 2D. These problems are challenging partly because of local minima generated by complex distributions of obstacles and other constraints such as velocity limits. The sensitivity of planners to parameter settings further adds to the difficulty. This is captured by the datasets we consider, where the $\mathtt{forest}$ distribution consists of small obstacles scattered around the workspace and requires the robot to squeeze through several narrow corridors and the $\mathtt{tarpit}$ distribution contains a small number of larger obstacles clumped together near the center of the workspace and requires the robot to avoid the cluster of obstacles entirely. It is challenging for a single planner with fixed parameters to solve problems from both distributions.

\subsection{Implementation details}

All our experiments and training are performed on a desktop with 8 Intel Core i7-7700K @ 4.20GHz CPUs, 32GB RAM and a 12GB NVIDIA Titan Xp. We consider a 2D point robot in a cluttered environment and planning is done in a state space $\bm\theta_i = [x, y, \dot{x}, \dot{y}]^T$. The robot is represented as a circle with radius $r$ centered on its center of mass and the environment is a binary occupancy grid. A Euclidean signed distance field is computed from the occupancy grid to evaluate distance to obstacles and check collisions. We utilize the same collision likelihood factor as GPMP2~\cite{Dong-RSS-16}, $\bm h(\bm{\theta}_i) = c(\bm{x}(\bm{\theta}_i))$, where $\bm{x}(\bm{\theta}_i) = [x, y]^T$ is the position coordinates of the center of mass and the hinge loss cost function $c$ is
\begin{equation}
c(\bm{x}) = \begin{cases}
-d(\bm{x}) + \epsilon & d(\bm{x}) \leq \epsilon \\
0 & otherwise
\end{cases}
\end{equation}
where $\epsilon = r + \epsilon_{safe}$ with $\epsilon_{safe}$ as a user defined safety distance, and $d$ is the signed distance.  In our current experiments, we consider $\sigma_{obs}$ as the learned parameter $\bm{\phi}_{L}$ and $\mathbf{Q}_C, \epsilon_{safe}, \bm{\mathcal{K}}_{s}, \bm{\mathcal{K}}_{g}$ to be the fixed parameters $\bm{\phi}_{F}$. Although, performance of the planner depends on both $\mathbf{Q}_C$ and $\mathbf{\Sigma}$, for our task they trade off against each other and thus we can achieve a similar behavior by varying one relative to the other. Since in our setup the environment changes, learning the likelihood covariance $\mathbf{\Sigma}$ is more relevant. In other problem domains learning $\mathbf{Q}_C$ instead might be more relevant such as~\cite{Rana-CoRL-17}. It is important to note the difference in expressiveness of $\bm{\Sigma}$ between GPMP2 where $\bm{\Sigma} = \sigma_{obs}^2 \times \mathbf{I}$, and \dplan where $\bm{\Sigma} = \mathrm{diag}(\sigma_{obs_1}^2, \ldots, \sigma_{obs_N}^2)$ with any $\sigma_{obs_i}$ being a function of the current trajectory and the environment.

\begin{figure}[!t]
	\centering
	\begin{subfigure}[t]{0.105\textwidth}
		\centering
		\includegraphics[trim={130 55 130 55},clip,width=\linewidth]{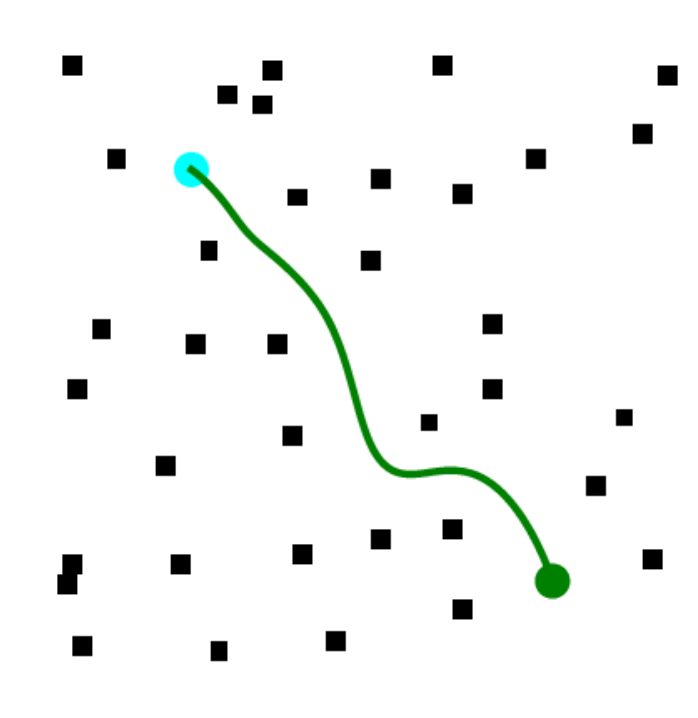}\vspace{-3mm}
		\includegraphics[trim={130 55 130 70},clip,width=\linewidth]{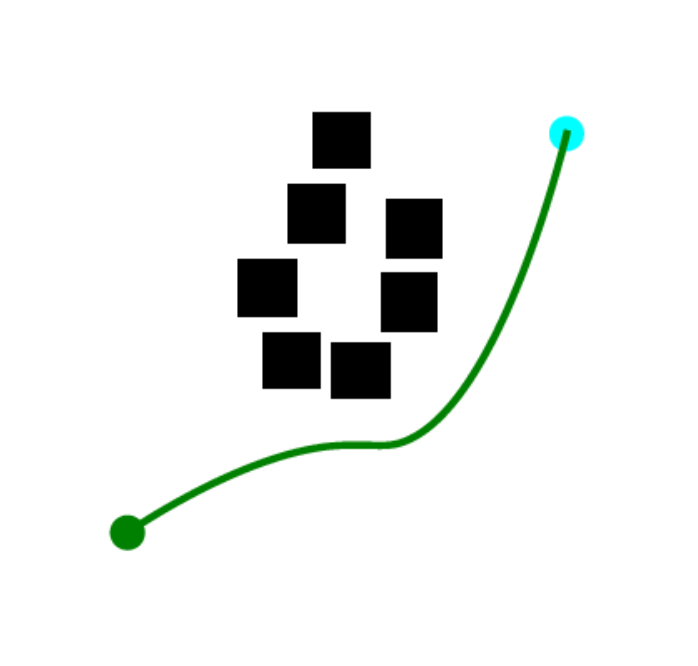}
		\caption{Expert}
		\label{fig:forest_expert}
	\end{subfigure}
	\hspace{1mm}
	\begin{subfigure}[t]{0.105\textwidth}
		\centering
		\includegraphics[trim={130 55 130 55},clip,width=\linewidth]{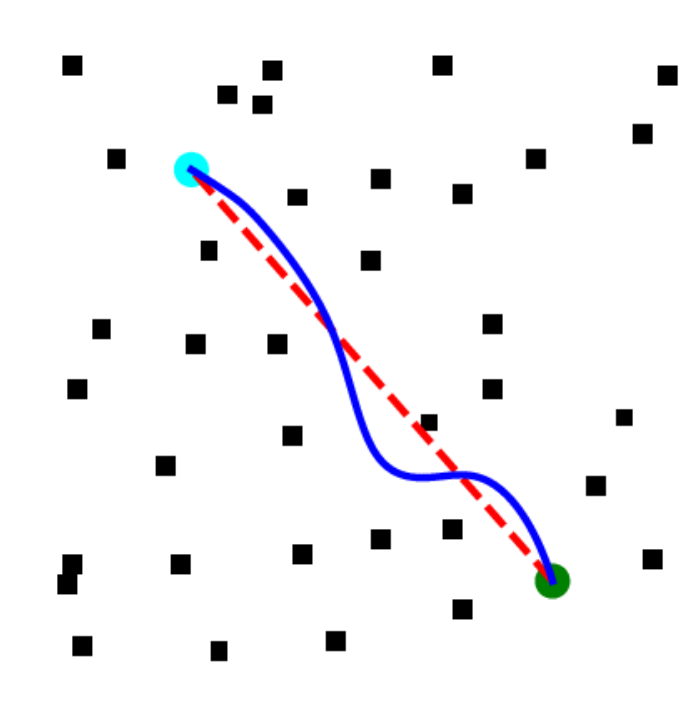}\vspace{-3mm}
		\includegraphics[trim={130 55 130 70},clip,width=\linewidth]{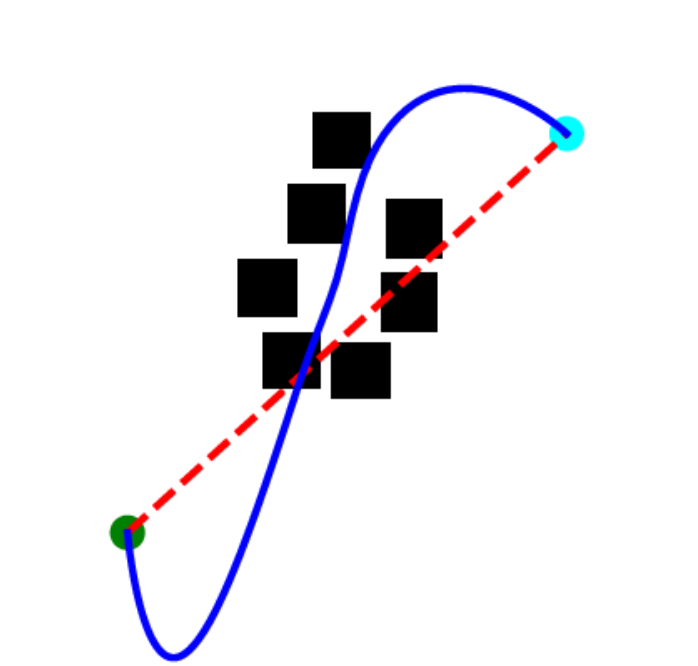}
		\caption{GPMP2,\\ ${\sigma}_{obs}=0.15$}
		\label{fig:exp_forest_015}
	\end{subfigure}
	\hspace{1mm}
	\begin{subfigure}[t]{0.105\textwidth}
		\centering
		\includegraphics[trim={130 55 130 55},clip,width=\linewidth]{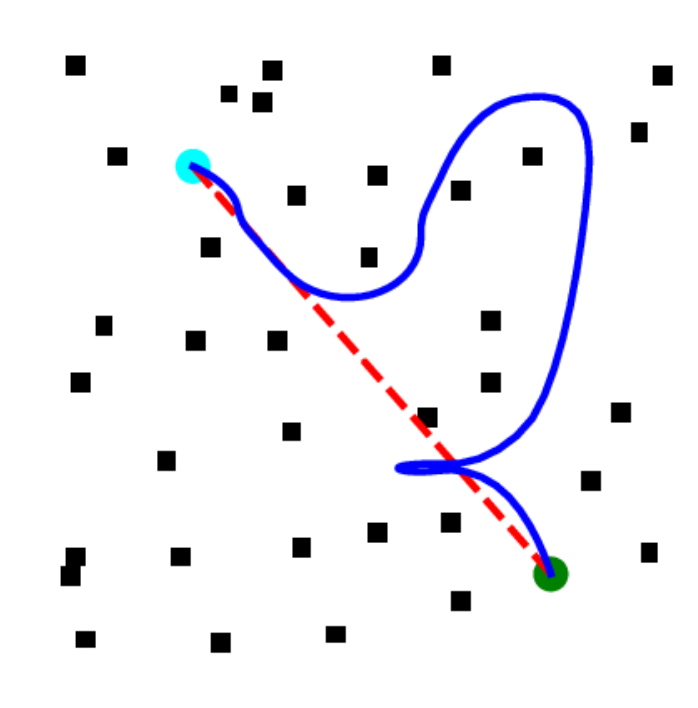}\vspace{-3mm}
		\includegraphics[trim={130 55 130 70},clip,width=\linewidth]{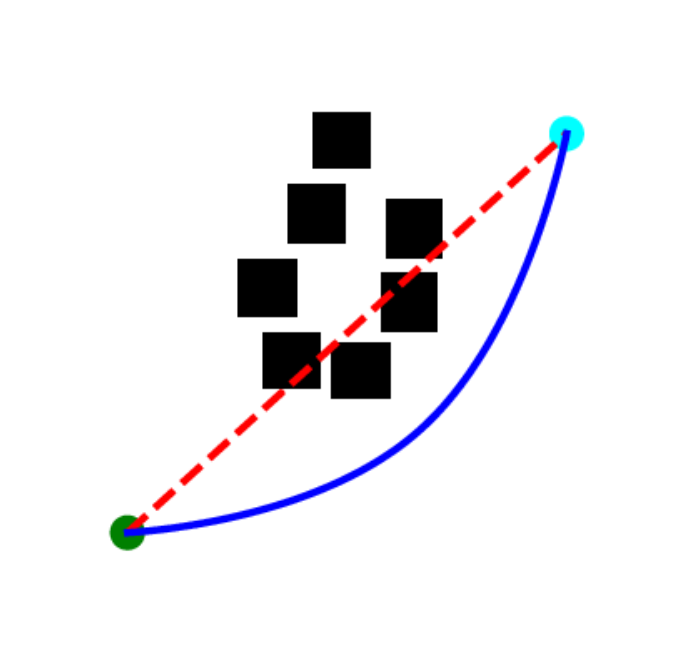}
		\caption{GPMP2,\\ ${\sigma}_{obs}=0.01$}
		\label{fig:exp_forest_001}
	\end{subfigure}
	\hspace{1mm}
	\begin{subfigure}[t]{0.105\textwidth}
		\centering
		\includegraphics[trim={130 55 130 55},clip,width=\linewidth]{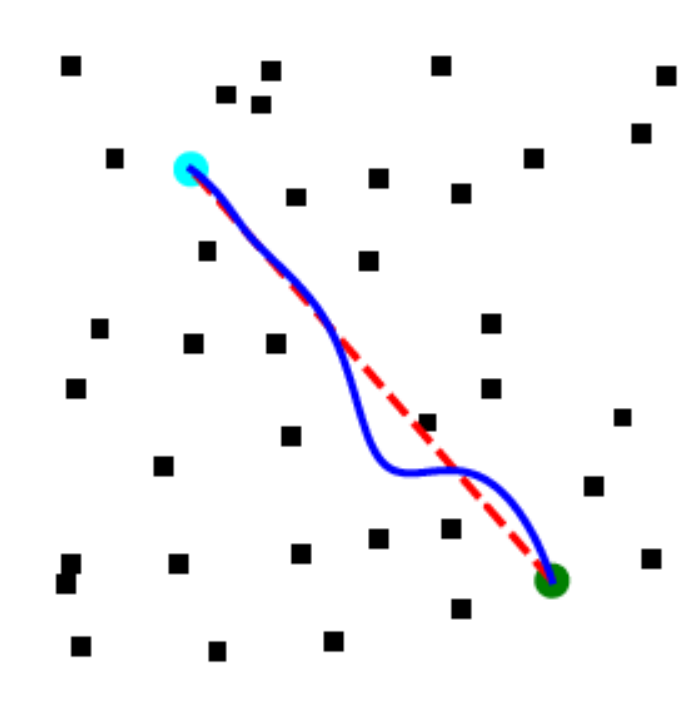}\vspace{-3mm}
		\includegraphics[trim={130 55 130 70},clip,width=\linewidth]{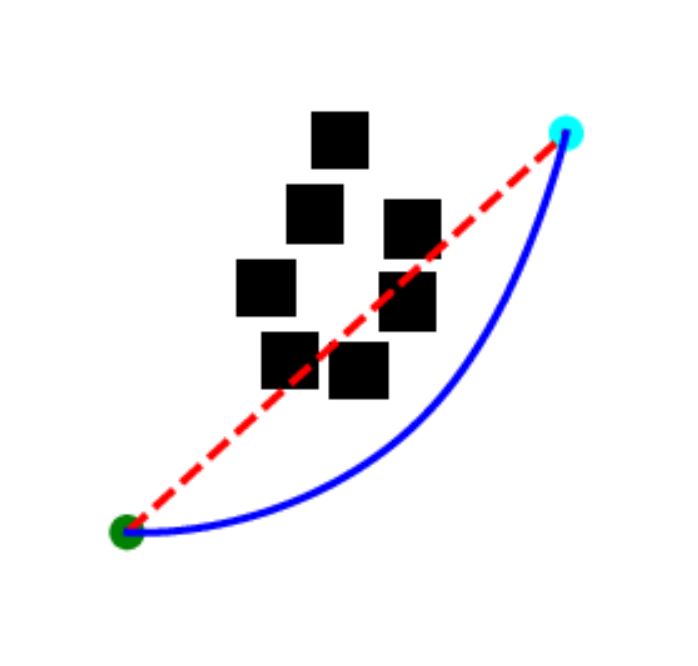}
		\caption{\dplan}
	\end{subfigure}
	\vspace{-1mm}
	\caption{\small Example comparison of (d) \dplan against (b)-(c) GPMP2 (fixed hand tuned covariances) and (a) Expert on $\mathtt{forest}$ (top row) and $\mathtt{tarpit}$ (bottom row) datasets. Hand tuned covariances that work well on one distribution of obstacles fail on the other and vice versa. By imitating the expert, \dplan is able to perform consistently across different environment distributions. Green circle is start, cyan is goal, dashed red line is initialization, and $\bm{Q}_{c} = 0.5 \times I$, $r = 0.4m$ for all. Trajectory is in collision if at any state the signed distance between robot center of mass and nearest obstacle is less than or equal to $r$.}
	\label{fig:comparision}
	\vspace{-8mm}
\end{figure}

\textbf{Loss function:}\quad
It can be expensive to gather a large number of human demonstrations to train the planner. Hence, we use a self-supervised approach.
Sampling based asymptotically optimal planning methods such as RRT*~\cite{karaman2011sampling} are effective in finding good homotopy classes to serve as an initialization for local trajectory optimizers, but can be slow to converge and produce non-smooth solution paths. We use a combination of RRT* and GPMP2 as our expert. Expert trajectories are generated by first running RRT* and are then optimized with GPMP2 to yield smooth solutions. This allows \dplan to learn by utilizing the best combination of local and global planning. We use the loss function defined in Section~\ref{sec:learn_cov} with this expert.

\textbf{Network architecture:}\quad
For $\mathbf{W}$ we use a standard feed-forward neural network model consisting of convolutional and fully connected layers. The network consists of 5 convolutional layers with [16, 16, 16, 32, 32] filters respectively, all 3x3 in size. This is followed by two fully connected layers with [1000, 640] hidden units. We use ReLU activation with batch normalization in all layers and a dropout probability of 0.5 in the fully connected layers. The input to the neural network is a 128x128 bitmap of the environment stacked on top of the euclidean signed distance field of the same dimensions. Backpropagation is performed for fixed number of iterations, $T = 10$. At every iteration, the network outputs a different likelihood covariance for each state along the trajectory.

\textbf{Comparing planners:}\quad
The convergence for the optimization is based on the following criterion: a tolerance on the relative change in error across iterations $\mathtt{tol}(\delta_{error})$, magnitude of update $\mathtt{tol}(\delta {\bm{\theta}})$, and max iterations $T_{max}$. On convergence the final trajectory is returned.  We report the following metrics on a test set of environments: (i) $\mathtt{success}$, percent of problems solved i.e. when a collision free trajectory is found, (ii) average $\mathtt{gp\_mse}$, mean-squared GP error measuring smoothness and (iii) $\mathtt{collision \_ intensity}$, the average portion of trajectory spent in collision when a collision occurs.

We test our framework on two different planning tasks to demonstrate (i) how learning covariances improves performance and (ii) how the planner's structure allows us to incorporate constraints. We compare against a baseline GPMP2 with hand-tuned parameters. However, we do not compare against other sampling and optimization-based planners and refer the reader to~\cite{Mukadam-IJRR-18} for benchmarks of GPMP2 against leading sampling and optimization-based planners.

\begin{table}
\centering
\caption{\small
Comparison of \dplan versus GPMP2 with fixed hand tuned covariances. \dplan learns the obstacle covariance $\sigma_{obs}$ using training set of 5000 environments. $\bm{Q}_{C} = 0.5 \times I$ for all. Total trajectory time is 10s with 100 states along the trajectory and $\lambda=1.0$ for training.}
\label{tab:mixed_clutter_num_solved}
\begin{tabular}{|r|r|c|c|c|}
\hline

\multicolumn{2}{|c|}{\multirow{2}{*}{}} & \multicolumn{2}{c|}{GPMP2} & \multirow{2}{*}{\dplan} \\
\multicolumn{2}{|c|}{} & \textbf{$\sigma_{obs} = 0.15$} & \textbf{$\sigma_{obs} = 0.01$} & \\
\hline
$\mathtt{forest}$ only & \multirow{3}{*}{$\mathtt{success}$} & \cmark{71.02} & 52.18 & 66.67 \\
$\mathtt{tarpit}$ only & & 55.56 & \cmark{74.08} & 68.00  \\
$\mathtt{mixed}$ & & 62.67 & 64.00  & \cmark{67.33}  \\
\hline
\multicolumn{2}{|r|}{$\mathtt{gp\_mse}$} & 0.002 & 0.0484 & \cmark{0.0015} \\
\multicolumn{2}{|r|}{$\mathtt{num\_iters}$} & 55.69 & 86.74 & \cmark{50.00}  \\
\multicolumn{2}{|r|}{$\mathtt{coll\_intensity}$} & 0.0464 &  0.0414 & \cmark{0.0374}  \\
\hline
\end{tabular}
\vspace{-5mm}
\end{table}

\subsection{Learning on complex distributions}
In this experiment, we show that if the planner's parameters are fixed, performance can be highly sensitive to distribution of obstacles in the environment. However, if a function can be learned to set the parameters  based on the current planning problem, this can help the planner achieve uniformly good performance across different obstacle distributions. We construct a hybrid dataset which is a mixture of two distinct distributions of obstacles, $\mathtt{forest}$ and $\mathtt{tarpit}$, as shown in Fig.~\ref{fig:comparision}. We use a test set of 150 randomly sampled environments from this mixed dataset and further subdivide it into two sets for each of the constituent distributions (roughly equal in proportion). We then hand-tuned parameters for GPMP2 to find the best covariances for the individual distributions and compared them against \dplan on three different test sets: two for the individual distributions and one for a mixed (roughly equal of the two distributions). 
Since there is no formal mathematical procedure for tuning parameters or even well-known heuristics, we rely on a manual line-search. Although this can likely be automated to find best static covariances for one given environment distribution, it is not practical when the environment distribution changes or when the parameters need to be adapted based on the location of the robot in the environment or the time-step on the trajectory.

The results in Table~\ref{tab:mixed_clutter_num_solved} show that for GPMP2 the best parameters on one distribution perform poorly on the other distribution in terms of success, although their performances on the mixed dataset are similar. Conversely, \dplan has uniform and consistent performance across both distributions even though it is only trained on the mixed dataset. This demonstrates that \dplan does not require manual tuning for every distribution of planning environments, but can automatically predict the covariances to use based on the current trajectory and environment as can be seen in Fig.~\ref{fig:comparision}. Additionally, \dplan has the lowest $\mathtt{gp\_mse}$ on the mixed dataset meaning the trajectories produced are still smooth. \dplan also converges in fewer number of iterations than the GPMP2 due to the covariance being more expressive and varying over iterations.

\textbf{Limitations:}\quad Since BPTT is known to have issues with exploding and vanishing gradients for long sequences, we use a small number of iterations ($T=10$) during training which prevents the learner from sufficiently exploring during training. The network architecture is a simple feed-forward network and does not have any memory and hence the learner does not learn to escape local minima very well. We believe that these issues can be addressed in the future using learning techniques such as Truncated Backpropagation Through Time (TBPTT)~\cite{williams1990efficient}, policy gradient methods~\cite{schulman2015trpo, schulman2015gae}, and recurrent networks such as LSTMs~\cite{Hochreiter1997}.

\subsection{Planning with velocity constraints}

We show that our learning method can explicitly incorporate planning constraints by including velocity limit factors into the optimization. We use a hinge loss similar to obstacle cost to bound the robot velocity $\dot{x}$ and $\dot{y}$ and set the covariance to a low value, $\bm{\mathcal{K}}_{v} = 10^{-4}$, analogous to joint limit factors in~\cite{Mukadam-IJRR-18}. We evaluate the average $\mathtt{constraint \_ violation}$ on a dataset with multiple randomly placed obstacles and study the effect of incorporating constraints during training. Table~\ref{tab:vel_constr_results} shows a comparison between \dplan trained with mild constraints and tested on problems with mild and tight constraints versus \dplan trained using tight constraints and tested on problems with tight constraints (details in the Table~\ref{tab:vel_constr_results} caption). We see that, by incorporating tight constraints during training, \dplan can learn to handle tight constraints while avoiding obstacles.   This illustrates that \dplan can successfully incorporate constraints within its structure, and that the method can learn to plan while respecting user-defined planning constraints.

\begin{table}
\centering
\caption{\small	Performance of \dplan with velocity constraints on different combinations of training and testing. Mild constraints are $v_{xmax} = 1.5m/s$, $v_{ymax} = 1.5m/s$, and $maxtime=15s$, tight constraints are $v_{xmax} = 1.0m/s$, $v_{ymax} = 1.0m/s$, and $maxtime=10s$ for the same start and goal. $maxtime$ is maximum time allowed for the trajectory.}
\begin{tabular}{|r|c|c|c|}
\hline
Training condition & {Mild} & {Mild} & {Tight} \\
Testing condition & {Mild} & {Tight} & {Tight} \\
\hline
$\mathtt{success}$ & 96 & 96 &98.12  \\
$\mathtt{constraint \_ violation}$ &0.0022 & 0.104 & 0.097 \\
\hline
\end{tabular}
\label{tab:vel_constr_results}
\vspace{-4mm}
\end{table}

\section{Discussion}\label{sec:discussion}

In this work, we developed  \dplan, a novel differentiable motion planning algorithm, by reformulating GPMP2 as a differentiable computational graph. Our method learned to predict objective function parameters as part of the differentiable planner and demonstrated competitive performance against planning with fixed, hand-tuned parameters.  
Our experimental results show that this strategy is an effective way to leverage experience to further improve upon traditional state-of-the-art motion planning algorithms.
We currently limited our experiments to only point robots in 2D environments to investigate the properties of the algorithm in a controlled setting. However, since the formulation was built on the GPMP2 planner, we believe that it can be extended to handle more complicated motion planning problems including articulated robots in 3D workspaces.


\bibliographystyle{IEEEtran}
\bibliography{IEEEabrv,refs}

\begin{thebibliography}{10}
\providecommand{\url}[1]{#1}
\csname url@samestyle\endcsname
\providecommand{\newblock}{\relax}
\providecommand{\bibinfo}[2]{#2}
\providecommand{\BIBentrySTDinterwordspacing}{\spaceskip=0pt\relax}
\providecommand{\BIBentryALTinterwordstretchfactor}{4}
\providecommand{\BIBentryALTinterwordspacing}{\spaceskip=\fontdimen2\font plus
\BIBentryALTinterwordstretchfactor\fontdimen3\font minus
  \fontdimen4\font\relax}
\providecommand{\BIBforeignlanguage}[2]{{%
\expandafter\ifx\csname l@#1\endcsname\relax
\typeout{** WARNING: IEEEtran.bst: No hyphenation pattern has been}%
\typeout{** loaded for the language `#1'. Using the pattern for}%
\typeout{** the default language instead.}%
\else
\language=\csname l@#1\endcsname
\fi
#2}}
\providecommand{\BIBdecl}{\relax}
\BIBdecl

\bibitem{ratliff2009chomp}
N.~Ratliff, M.~Zucker, J.~A. Bagnell, and S.~Srinivasa, ``Chomp: Gradient
  optimization techniques for efficient motion planning,'' in \emph{ICRA},
  2009.

\bibitem{schulman2013finding}
J.~Schulman, J.~Ho, A.~Lee, I.~Awwal, H.~Bradlow, and P.~Abbeel, ``Finding
  locally optimal, collision-free trajectories with sequential convex
  optimization.'' in \emph{RSS}, 2013.

\bibitem{Mukadam-IJRR-18}
M.~Mukadam, J.~Dong, X.~Yan, F.~Dellaert, and B.~Boots, ``Continuous-time
  {G}aussian process motion planning via probabilistic inference,'' \emph{The
  International Journal of Robotics Research (IJRR)}, vol.~37, no.~11, pp.
  1319--1340, 2018.

\bibitem{Dong-RSS-16}
J.~Dong, M.~Mukadam, F.~Dellaert, and B.~Boots, ``Motion planning as
  probabilistic inference using {G}aussian processes and factor graphs,'' in
  \emph{Proceedings of Robotics: Science and Systems (RSS-2016)}, 2016.

\bibitem{faust2018prm}
A.~Faust, K.~Oslund, O.~Ramirez, A.~Francis, L.~Tapia, M.~Fiser, and
  J.~Davidson, ``{PRM-RL}: Long-range robotic navigation tasks by combining
  reinforcement learning and sampling-based planning,'' in \emph{2018 IEEE
  International Conference on Robotics and Automation (ICRA)}.\hskip 1em plus
  0.5em minus 0.4em\relax IEEE, 2018, pp. 5113--5120.

\bibitem{ratliff2009learning}
N.~D. Ratliff, D.~Silver, and J.~A. Bagnell, ``Learning to search: Functional
  gradient techniques for imitation learning,'' \emph{Autonomous Robots},
  vol.~27, no.~1, pp. 25--53, 2009.

\bibitem{bhardwaj2017heuristic}
M.~Bhardwaj, S.~Choudhury, and S.~Scherer, ``Learning heuristic search via
  imitation,'' in \emph{Conference on Robot Learning}, 2017, pp. 271--280.

\bibitem{Bhardwaj19a}
M.~Bhardwaj, S.~Choudhury, B.~Boots, and S.~Srinivasa, ``Leveraging experience
  in lazy search,'' 2019.

\bibitem{tamar2016value}
A.~Tamar, Y.~Wu, G.~Thomas, S.~Levine, and P.~Abbeel, ``Value iteration
  networks,'' in \emph{Advances in Neural Information Processing Systems},
  2016, pp. 2154--2162.

\bibitem{srinivas2018universal}
A.~Srinivas, A.~Jabri, P.~Abbeel, S.~Levine, and C.~Finn, ``Universal planning
  networks: Learning generalizable representations for visuomotor control,'' in
  \emph{International Conference on Machine Learning}, 2018, pp. 4739--4748.

\bibitem{clark2018learning}
R.~Clark, M.~Bloesch, J.~Czarnowski, S.~Leutenegger, and A.~J. Davison,
  ``Learning to solve nonlinear least squares for monocular stereo,'' in
  \emph{Proceedings of the European Conference on Computer Vision (ECCV)},
  2018, pp. 284--299.

\bibitem{NIPS2016_6461}
M.~Andrychowicz, M.~Denil, S.~G\'{o}mez, M.~W. Hoffman, D.~Pfau, T.~Schaul,
  B.~Shillingford, and N.~de~Freitas, ``Learning to learn by gradient descent
  by gradient descent,'' in \emph{Advances in Neural Information Processing
  Systems 29}, 2016, pp. 3981--3989.

\bibitem{chen2018neural}
T.~Q. Chen, Y.~Rubanova, J.~Bettencourt, and D.~K. Duvenaud, ``Neural ordinary
  differential equations,'' in \emph{Advances in Neural Information Processing
  Systems}, 2018, pp. 6572--6583.

\bibitem{amos2017optnet}
B.~Amos and J.~Z. Kolter, ``Optnet: Differentiable optimization as a layer in
  neural networks,'' in \emph{Proceedings of the 34th International Conference
  on Machine Learning-Volume 70}.\hskip 1em plus 0.5em minus 0.4em\relax JMLR.
  org, 2017, pp. 136--145.

\bibitem{amos2018differentiable}
B.~Amos, I.~Jimenez, J.~Sacks, B.~Boots, and J.~Z. Kolter, ``Differentiable mpc
  for end-to-end planning and control,'' in \emph{Advances in Neural
  Information Processing Systems}, 2018, pp. 8299--8310.

\bibitem{byravan2018se3}
A.~Byravan, F.~Lceb, F.~Meier, and D.~Fox, ``Se3-pose-nets: Structured deep
  dynamics models for visuomotor control,'' in \emph{2018 IEEE International
  Conference on Robotics and Automation (ICRA)}.\hskip 1em plus 0.5em minus
  0.4em\relax IEEE, 2018, pp. 1--8.

\bibitem{cano2018automatic}
J.~Cano, Y.~Yang, B.~Bodin, V.~Nagarajan, and M.~O'Boyle, ``Automatic parameter
  tuning of motion planning algorithms,'' in \emph{2018 IEEE/RSJ International
  Conference on Intelligent Robots and Systems (IROS)}.\hskip 1em plus 0.5em
  minus 0.4em\relax IEEE, 2018, pp. 8103--8109.

\bibitem{burger2017automated}
R.~Burger, M.~Bharatheesha, M.~van Eert, and R.~Babu{\v{s}}ka, ``Automated
  tuning and configuration of path planning algorithms,'' in \emph{2017 IEEE
  International Conference on Robotics and Automation (ICRA)}.\hskip 1em plus
  0.5em minus 0.4em\relax IEEE, 2017, pp. 4371--4376.

\bibitem{barfoot2014batch}
T.~D. Barfoot, C.~H. Tong, and S.~S{\"a}rkk{\"a}, ``Batch continuous-time
  trajectory estimation as exactly sparse gaussian process regression.''\hskip
  1em plus 0.5em minus 0.4em\relax Citeseer, 2014.

\bibitem{paszke2017automatic}
A.~Paszke, S.~Gross, S.~Chintala, G.~Chanan, E.~Yang, Z.~DeVito, Z.~Lin,
  A.~Desmaison, L.~Antiga, and A.~Lerer, ``Automatic differentiation in
  pytorch,'' 2017.

\bibitem{dellaert2006square}
F.~Dellaert and M.~Kaess, ``Square root {SAM}: Simultaneous localization and
  mapping via square root information smoothing,'' \emph{The International
  Journal of Robotics Research}, vol.~25, no.~12, pp. 1181--1203, 2006.

\bibitem{karaman2011sampling}
S.~Karaman and E.~Frazzoli, ``Sampling-based algorithms for optimal motion
  planning,'' \emph{The International Journal of Robotics Research}, vol.~30,
  no.~7, pp. 846--894, 2011.

\bibitem{Rana-CoRL-17}
M.~A. Rana, M.~Mukadam, S.~R. Ahmadzadeh, S.~Chernova, and B.~Boots., ``Towards
  robust skill generalization: Unifying learning from demonstration and motion
  planning,'' in \emph{Proceedings of the 2017 Conference on Robot Learning
  ({CoRL})}, 2017.

\bibitem{zucker2013chomp}
M.~Zucker, N.~Ratliff, A.~D. Dragan, M.~Pivtoraiko, M.~Klingensmith, C.~M.
  Dellin, J.~A. Bagnell, and S.~S. Srinivasa, ``Chomp: Covariant hamiltonian
  optimization for motion planning,'' \emph{The International Journal of
  Robotics Research}, vol.~32, no. 9-10, pp. 1164--1193, 2013.

\bibitem{rumelhart1988learning}
D.~E. Rumelhart, G.~E. Hinton, R.~J. Williams \emph{et~al.}, ``Learning
  representations by back-propagating errors,'' \emph{Cognitive modeling},
  vol.~5, no.~3, p.~1, 1988.

\bibitem{williams1990efficient}
R.~J. Williams and J.~Peng, ``An efficient gradient-based algorithm for on-line
  training of recurrent network trajectories,'' \emph{Neural computation},
  vol.~2, no.~4, pp. 490--501, 1990.

\bibitem{schulman2015trpo}
J.~Schulman, S.~Levine, P.~Abbeel, M.~Jordan, and P.~Moritz, ``Trust region
  policy optimization,'' in \emph{ICML}, 2015.

\bibitem{schulman2015gae}
J.~Schulman, P.~Moritz, S.~Levine, M.~Jordan, and P.~Abbeel, ``High-dimensional
  continuous control using generalized advantage estimation,'' \emph{arXiv
  preprint arXiv:1506.02438}, 2015.

\bibitem{Hochreiter1997}
S.~Hochreiter and J.~Schmidhuber, ``Long short-term memory,'' \emph{Neural
  Comput.}, vol.~9, no.~8, pp. 1735--1780, Nov. 1997.

\end{thebibliography}


\end{document}